\documentclass[11pt,a4paper]{article}
\usepackage[hyperref]{emnlp2020}
\usepackage{times}
\usepackage{latexsym}
\usepackage{graphicx}
\usepackage{subfig}
\usepackage{array}
\usepackage{xcolor}
\usepackage{colortbl}

\usepackage{url}

\usepackage{amssymb}

\aclfinalcopy 

\title{LemMED: Fast and Effective Neural Morphological Analysis with Short Context Windows}

\author{
  Aibek Makazhanov, Sharon Goldwater, Adam Lopez \\

  School of Informatics \\
  University of Edinburgh \\

  {\tt aibek.makazhanov@ed.ac.uk} \\
  {\tt sgwater@inf.ed.ac.uk},
  {\tt alopez@inf.ed.ac.uk}
  \\
}

\date{}

\begin{document}
\maketitle
\begin{abstract}

We present LemMED, a character-level encoder-decoder for contextual morphological analysis (combined lemmatization and tagging).
LemMED extends and is named after two other attention-based models, namely Lematus, a contextual lemmatizer, and MED, a morphological (re)inflection model.
Our approach does not require training separate lemmatization and tagging models, nor does it need additional resources and tools, such as morphological dictionaries or transducers.
Moreover, LemMED relies solely on character-level representations and on local context.
Although the model can, in principle, account for global context on sentence level, our experiments show that using just a single word of context around each target word is not only more computationally feasible, but yields better results as well.
We evaluate LemMED in the framework of the SIMGMORPHON-2019 shared task on combined lemmatization and tagging.
In terms of average performance LemMED ranks 5th among 13 systems and is bested only by the submissions that use contextualized embeddings.
\end{abstract}

\section{\label{INTRO} Introduction}

Contextual morphological analysis is the task of assigning each surface form exactly one lexical form that is most probable in the given context.
Here, \textit{surface form} is a token of running text and \textit{lexical form} (or \textit{analysis}) is a tuple containing a lemma and a morphosyntactic description, i.e. morpho-tag\footnote{We define a morpho-tag as a list of values of grammatical categories (grammemes) present in the given surface form, e.g. [noun, plural] for \textit{cats}. In this work POS tags are considered parts of morpho-tags and the terms \textit{morpho-tag}, \textit{tag}, and \textit{morphosyntactic description} are used interchangeably.}. 
For example, a possible output for \textit{``bats bit cats"} could be [\textit{bat}; n, pl] [\textit{bite}; v, pst, 3pl] [\textit{cat}; n, pl], where lexical forms are given in brackets with lemmata emphasized in italics.



Most of the existing work employs one of two approaches, modeling either (i) a distribution of 
lexical forms\footnote{Typically lexical forms are distributed much sparser than their constituents (tags and, to a lesser degree, lemmata). Therefore actual practical implementations model marginal distributions of either only tags~\cite{hakkani2002statistical} or tags and (parts of) lemmata~\cite{STRAKA16.873}.} 
(as whole units) or (ii) a joint distribution of lemmata and tags.
The first approach is commonly referred to as morphological disambiguation, as the problem is reduced to choosing the most probable analysis from a list of candidates, generated for each surface form either with the help of finite-state transducers~\cite{Koskenniemi1983TwoLevelMF, WASHINGTON14.1207} or by means of various heuristics such as trie-based dictionary induction~\cite{STRAKA16.873} or morpheme transition trees~\cite{disam_kazakh}.
The second approach avoids using FST or heuristics by either training separate lemmatization and tagging models~\cite{morfette, Lemming} or resorting to multi-task learning~\cite{sgm2019-UFAL, sgm2019-CSAAR}.

While both of the aforementioned approaches cast the problem as a sequence labeling task, recent work has explored various sequence-to-sequence modeling strategies, where either a lemma~\cite{kondratyuk2018lemmatag, neubsl} or a tag~\cite{sgm2019-OSU} or both~\cite{morphnet18, akyurek2019, sgm2019-ITU, sgm2019-RUG} are generated one character/grammeme at a time by means of neural encoder-decoder architecture.
We follow the same strategy, but unlike listed work, rely solely on character-level context and enable the decoder to generate more than one analysis at a time.
In this respect our approach can be viewed as an extension of character-level attention models developed for the related tasks, namely MED, morphological (re)inflection model~\cite{medACL}, and Lematus, contextual lemmatizer~\cite{LEMATUS}.




We present LemMED, a character-level encoder-decoder with attention that requires neither training separate lemmatization and tagging models, nor using external dictionaries or transducers.
Given a corpus annotated for morpho-syntactic information, LemMED learns to directly model a sequence of lexical forms from a given sequence of surface forms.
This is done in two modes.
In the full sequence mode, LemMED analyzes an entire sentence at once and in the context window mode, LemMED learns to model a sequence of lexical forms for each input token and surrounding context words.

In our experiments we show that in the context window mode LemMED is both more accurate and efficient than in the full sequence mode.
We evaluate LemMED on a set of more than a hundred treebanks, comparing it to both popular baselines available off-the-shelf and state of the art systems submitted to the SIGMORPHON-2019 shared task on contextual analysis.
In both cases our model achieves competitive performance and when compared to the baselines shows clear dominance in terms of combined lemmatization and tagging accuracy and performance on unseen data.

\section{\label{MODEL} LemMED}

\begin{figure*}[!t]


\subfloat[Contextual analysis by LemMED in the full sequence mode.]{%
  \scriptsize
  \begin{tabular}{r|c|c|c|c|c|c|c|c|c|c|c|c|c|c|c|c|c|c|c|cccc}
    \cline{2-20}
    \texttt{source:} & 
    \texttt{B} & \texttt{a} & \texttt{t} & \texttt{s} & 
    \multicolumn{2}{r|}{\texttt{\#}} & 
    \texttt{b} & \texttt{i} & \texttt{t} & 
    \multicolumn{4}{r|}{\texttt{\#}} & 
    \texttt{c} & \texttt{a} & \texttt{t} & \texttt{s} & 
    \multicolumn{2}{r|}{\texttt{\#}} \\
    \cline{2-20}
    \multicolumn{20}{c} ~ \vspace{-5pt} \\
    \cline{2-20}
    \texttt{target:} & 
    \texttt{b} & \texttt{a} & \texttt{t} & 
    \texttt{+n} & \texttt{+pl} & \texttt{\#} & 
    \texttt{b} & \texttt{i} & \texttt{t} & \texttt{e} & 
    \texttt{+v} & \texttt{+pst} & \texttt{\#} & 
    \texttt{c} & \texttt{a} & \texttt{t} & \texttt{+n} & 
    \texttt{+pl} & \texttt{\#} & \texttt{~} & \texttt{~} & \texttt{~} \\
    \cline{2-20}
    \multicolumn{20}{c} ~ \vspace{-7pt} \\
  \end{tabular}
  \label{EXFIG-LemMED-FS}
}
\vspace{2pt}

\subfloat[Non-contextual inflection by MED. Vertical lines signify isolated processing of each source-target pair.]{%
  \scriptsize
  \begin{tabular}{r|c|c|c|c|c|c|c|c|c|c|c|c|c|c|c|c|c|c|c|c|ccc}



    

    


    

    \cline{2-6}\cline{9-14}\cline{17-21}
    \texttt{source:} & 
    \texttt{b} & \texttt{a} & \texttt{t} & \texttt{+n} & 
    \texttt{+pl} & 
    ~ & ~ &
    
    \texttt{b} & \texttt{i} & \texttt{t} & \texttt{e} & 
    \texttt{+v} & \texttt{+pst} & 
    ~ & ~ & 
    
    \texttt{c} & \texttt{a} & \texttt{t} & \texttt{+n} & 
    \texttt{+pl} & ~ & ~ & ~ \\
    
    \cline{2-6}\cline{9-14}\cline{17-21}
    \multicolumn{7}{c|}{\vspace{-5pt}} & 
    \multicolumn{8}{c|}{~} \\
    \cline{2-5}\cline{9-11}\cline{17-20}
    
    \texttt{target:} & 
    \texttt{b} & \texttt{a} & \texttt{t} & \texttt{s} & 
    \multicolumn{2}{c|}{~} & ~ & 
    
    \texttt{b} & \texttt{i} & \texttt{t} & 
    \multicolumn{4}{c|}{~} & ~ & 
    \texttt{c} & \texttt{a} & \texttt{t} & \texttt{s} \\
    
    \cline{2-5}\cline{9-11}\cline{17-20}
    \multicolumn{4}{c} ~ \vspace{-7pt} \\
  \end{tabular}
  \label{EXFIG-MED}
}
\vspace{2pt}

\subfloat[
Contextual lemmatization by Lematus with a seven-character context window.
The focal tokens are highlighted.
]{%
  \scriptsize
  \begin{tabular}{r|c|c|c|c|c|c|c|c|c|c|c|c|c|c|c|c|c|c|c|c|c|c|c|c|c|c|c|c|c|c}
    \cline{2-12}
    \texttt{source:} & 
    \cellcolor[gray]{0.79} \texttt{B} & 
    \cellcolor[gray]{0.79} \texttt{a} & 
    \cellcolor[gray]{0.79} \texttt{t} & 
    \cellcolor[gray]{0.79} \texttt{s} & \texttt{\#} & 
    \texttt{b} & \texttt{i} & \texttt{t} & \texttt{\#} &
    \texttt{c} & \texttt{a} \\
    \cline{2-12}
    \multicolumn{1}{c}{~ \vspace{-5pt}} \\
    \cline{2-4}
    \texttt{target:} & 
    \texttt{b} & \texttt{a} & \texttt{t} \\
    \cline{2-4}
    \multicolumn{1}{c}{~} \\

    \cline{2-15}
    \texttt{source:} & 
    \texttt{B} & \texttt{a} & \texttt{t} & \texttt{s} & \texttt{\#} & 
    \cellcolor[gray]{0.79} \texttt{b} & 
    \cellcolor[gray]{0.79} \texttt{i} & 
    \cellcolor[gray]{0.79} \texttt{t} & 
    \texttt{\#} &
    \texttt{c} & \texttt{a} & \texttt{t} & \texttt{s} & \texttt{\#} & 
    \multicolumn{1}{c}{~} & \multicolumn{1}{c}{~} & 
    \multicolumn{1}{c}{~} & \multicolumn{1}{c}{~} & 
    \multicolumn{1}{c}{~} & \multicolumn{1}{c}{~} & 
    \multicolumn{1}{c}{~} & \multicolumn{1}{c}{~} & 
    \multicolumn{1}{c}{~} & \multicolumn{1}{c}{~} & 
    \multicolumn{1}{c}{~} & \multicolumn{1}{c}{~} \\
    \cline{2-15}
    \multicolumn{1}{c}{~ \vspace{-5pt}} \\
    \cline{7-10}
    \multicolumn{1}{r}{\texttt{target:}} & 
    \multicolumn{5}{r|}{~} & 
    \texttt{b} & \texttt{i} & \texttt{t} & \texttt{e} \\
    \cline{7-10}
    \multicolumn{1}{c}{~} \\

    \cline{4-15}
    \multicolumn{1}{r}{\texttt{source:}} & 
    \multicolumn{2}{r|}{~} & 
    \texttt{t} & \texttt{s} & \texttt{\#} & 
    \texttt{b} & \texttt{i} & \texttt{t} & \texttt{\#} &
    \cellcolor[gray]{0.79} \texttt{c} & 
    \cellcolor[gray]{0.79} \texttt{a} & 
    \cellcolor[gray]{0.79} \texttt{t} & 
    \cellcolor[gray]{0.79} \texttt{s} & \texttt{\#} \\
    \cline{4-15}
    \multicolumn{1}{c}{~ \vspace{-5pt}} \\
    \cline{11-13}
    \multicolumn{1}{r}{\texttt{target:}} & 
    \multicolumn{9}{r|}{~} & 
    \texttt{c} & \texttt{a} & \texttt{t} \\
    \cline{11-13}
    \multicolumn{1}{c}{~} \\



  \end{tabular}


    
    
    
    
    

  \label{EXFIG-Lematus}
}

\subfloat[
Contextual analysis by LemMED with a single-word context window. 
For each source-target pair the highlighted parts denote the focal token and the corresponding lexical form.
]{%
  \scriptsize
  \begin{tabular}{r|c|c|c|c|c|r|c|c|c|c|c|c|c|c|c|c|c|c|c|c}
    
    \cline{2-14}
    \texttt{source:} & 
    \cellcolor[gray]{0.79} \texttt{B} & 
    \cellcolor[gray]{0.79} \texttt{a} & 
    \cellcolor[gray]{0.79} \texttt{t} & 
    \cellcolor[gray]{0.79} \texttt{s} & 
    \multicolumn{2}{r|}{\texttt{\#}} & 
    \texttt{b} & \texttt{i} & \texttt{t} & 
    \multicolumn{4}{r|}{\texttt{\#}} \\
    \cline{2-14}
    \multicolumn{20}{c} ~ \vspace{-5pt} \\
    \cline{2-14}
    \texttt{target:} & 
    \cellcolor[gray]{0.79} \texttt{b} & 
    \cellcolor[gray]{0.79} \texttt{a} & 
    \cellcolor[gray]{0.79} \texttt{t} & 
    \cellcolor[gray]{0.79} \texttt{+n} & 
    \cellcolor[gray]{0.79} \texttt{+pl} & 
    \texttt{\#} & 
    \texttt{b} & \texttt{i} & \texttt{t} & \texttt{e} & 
    \texttt{\textit{+v}} & \texttt{\textit{+pst}} & 
    \texttt{\#} \\
    \cline{2-14}
    \multicolumn{20}{c}{~} \\
    
    \cline{2-20}
    \texttt{source:} & 
    \texttt{B} & \texttt{a} & \texttt{t} & \texttt{s} & 
    \multicolumn{2}{r|}{\texttt{\#}} & 
    \cellcolor[gray]{0.79} \texttt{b} & 
    \cellcolor[gray]{0.79} \texttt{i} & 
    \cellcolor[gray]{0.79} \texttt{t} & 
    \multicolumn{4}{r|}{\texttt{\#}} & 
    \texttt{c} & \texttt{a} & \texttt{t} & \texttt{s} & 
    \multicolumn{2}{r|}{\texttt{\#}} \\
    \cline{2-20}
    \multicolumn{20}{c} ~ \vspace{-5pt} \\
    \cline{2-20}
    \texttt{target:} & 
    \texttt{b} & \texttt{a} & \texttt{t} & 
    \texttt{\textit{+n}} & \texttt{\textit{+pl}} & 
    \texttt{\#} & 
    \cellcolor[gray]{0.79} \texttt{b} & 
    \cellcolor[gray]{0.79} \texttt{i} & 
    \cellcolor[gray]{0.79} \texttt{t} & 
    \cellcolor[gray]{0.79} \texttt{e} & 
    \cellcolor[gray]{0.79} \texttt{+v} & 
    \cellcolor[gray]{0.79} \texttt{+pst} & 
    \texttt{\#} & 
    \texttt{c} & \texttt{a} & \texttt{t} & 
    \texttt{\textit{+n}} & \texttt{\textit{+pl}} & 
    \texttt{\#} & \texttt{~} \\
    \cline{2-20}
    \multicolumn{20}{c}{~} \\

    \cline{8-20}
    \multicolumn{1}{r}{\texttt{source:}} & 
    \multicolumn{6}{r|}{~} &
    \texttt{b} & \texttt{i} & \texttt{t} & 
    \multicolumn{4}{r|}{\texttt{\#}} & 
    \cellcolor[gray]{0.79} \texttt{c} & 
    \cellcolor[gray]{0.79} \texttt{a} & 
    \cellcolor[gray]{0.79} \texttt{t} & 
    \cellcolor[gray]{0.79} \texttt{s} & 
    \multicolumn{2}{r|}{\texttt{\#}} \\
    \cline{8-20}
    \multicolumn{20}{c} ~ \vspace{-5pt} \\
    \cline{8-20}
    \multicolumn{1}{r}{\texttt{target:}} & 
    \multicolumn{6}{r|}{~} &
    \texttt{b} & \texttt{i} & \texttt{t} & \texttt{e} & 
    \texttt{\textit{+v}} & \texttt{\textit{+pst}} & 
    \texttt{\#} & 
    \cellcolor[gray]{0.79} \texttt{c} & 
    \cellcolor[gray]{0.79} \texttt{a} & 
    \cellcolor[gray]{0.79} \texttt{t} & 
    \cellcolor[gray]{0.79} \texttt{+n} & 
    \cellcolor[gray]{0.79} \texttt{+pl} & \texttt{\#} \\
    \cline{8-20}
    \multicolumn{20}{c} ~ \vspace{-7pt} \\
  \end{tabular}
  \label{EXFIG-LemMED-CW}
}

\caption{\label{EXFIG} An example sentence and the corresponding sequence of analyses/lemmata as represented by LemMED, MED, and Lematus for their respective tasks.
The ``source" and ``target" labels convey the meaning generally attached to these terms in the sequence-to-sequence modeling framework and denote corresponding sequences.
In every such sequence each individual cell represents a single processing unit: either a character or a grammeme.
Hash sign (\#) denotes a word boundary\footnotemark.
}

\end{figure*}

LemMED is a character-level encoder-decoder with attention.
It is implemented in \mbox{OpenNMT-py} ~\cite{opennmt}, a Python-based open source toolkit designed for working with sequence-to-sequence models.
OpenNMT-py allows to assemble models from various pre-implemented components and to fine tune a rich set of parameters.
In our model we use a stacked bidirectional LSTM for the encoder, an LSTM~\cite{schuster1997blstm} for the decoder, and global attention~\cite{global_att}.
That is LemMED in a nutshell.
Despite the difference in the choice of encoder/decoder architectures, conceptually it is the same basic attention model, as the one used in Lematus~\cite{LEMATUS} for contextual lemmatization and in MED~\cite{medACL} for non-contextual inflection.
The only thing required to extend those approaches to the task at hand is to change what goes into source and target sequences that the models operate with.

LemMED is trained in a machine translation fashion, but instead of an aligned pair of sentences in two different languages, a training example contains the same sentence at the surface and lexical level (cf. Figure~\ref{EXFIG-LemMED-FS}).
If we now consider each pair of surface and lexical forms in isolation and flip the source and target sides, as shown on Figure~\ref{EXFIG-MED}, we get a representation used by MED for the inflection sub-task of the SIGMORPHON-2016 shared task~\cite{SIGM-ST16}.
Thus, in the full sequence mode, LemMED extends MED simply by accounting for context and changing the direction of processing from inflection to analysis.

Similarly to MED, Lematus treats each token-lemma pair separately, but not in isolation from the rest of the sentence.
Local context is retained on the source side in a form of surrounding characters captured by a fixed-sized context window (cf. Figure~\ref{EXFIG-Lematus}).
The context window is slid over a given sequence of surface forms and corresponding lemmata one form-lemma pair at a time.
This results in a representation that has a pair of short source and target sequences per token, rather than one pair of sequences per whole sentence.
We refer to this method of slicing the input into short snippets as ``the context window mode".
As can be seen from Figure~\ref{EXFIG-LemMED-CW}, in this mode our model extends Lematus by introducing three modifications.
First, the context window size is measured in words, not characters.
Second, as required by the task at hand, target sequences consist of complete lexical forms, not just lemmata.
Third, unlike Lematus that has one target lemma per snippet, our model has as many target-side forms as there are corresponding source-side forms.
Let us elaborate on the latter distinction.

By default, in the context window mode there is an equal number of surface and lexical forms on the source and target sides respectively.
This means that the model learns to predict target-side context (cf. Figure~\ref{EXFIG-LemMED-CW}, unhighlighted) along with the focal lexical form (highlighted).
Yet, at test time only the latter is considered for the final prediction and evaluation, while context is ignored.
Thus, including target-side context seems useless, as most of what is predicted gets thrown away, suggesting that the model optimizes for potentially redundant data.
On the other hand, this way we are able to implicitly incorporate lexical context, without making any changes to the architecture of the model.
Moreover, precisely because it is not evaluated, we can use virtually anything for target-side context, e.g. only lemmata or only tags or neither.
At any rate, particular usage of target-side context is more of a design choice, and various possibilities are compared in subsection~\ref{TGTSEQ}.
Here it must be noted that such flexibility is not available in the full sequence mode, as input tokens are considered all at once and lack their own target context\footnotetext{This symbol is used in our example only for presentation clarity.
In practice, to tell a word boundary from a running character, both Lematus and LemMED use special multi-character labels.}.





Finally, let us point out another important aspect of modeling target-side context.
Suppose that all three target sequences depicted on Figure~\ref{EXFIG-LemMED-CW} are the outputs generated by LemMED at test time.
Notice that each input token is predicted across multiple snippets.
Specifically, predictions for ``\textit{Bats}" and ``\textit{cats}" appear in the snippets \#1, 2 and \#2, 3 respectively, while predictions for ``\textit{bit}" appear in all of the snippets.
In general, given a context window of size $W$, sentence-initial and sentence-final tokens will be predicted $2W$ times, while predictions for the rest of the tokens will be made $2W+1$ times.
This allows for a majority voting scheme to be implemented, where for each input token the most frequent output is considered for the final prediction and evaluation.
Hence, if modeled, not only target side context may provide contextual clues during training, it can, after all, be used directly at test time.


\section{\label{ES} Experimental Setup}


\subsection{\label{DATA} Data Set}

We work with the SIGMORPHON-2019 shared task data~\cite{sgm2019}, sub-task \#2 (hereinafter Shared task).
The original data set consists of 107 UD treebanks~\cite{UD} converted to the Universal Morphology scheme~\cite{UM}.
We had to exclude two of the treebanks (Estonian-EDT and Portuguese-GSD), as one of our baselines (Lemming) kept crashing when trained on them.
Thus, our data set contains 105 treebanks that cover 65 different languages.

Table~\ref{tab_data} summarizes basic stats of the data set averaged over all treebanks and broken down into splits.
Notice how standard deviation values of sentence and token counts are more than 1.5 times larger than the respective means.
This suggest great variation in treebank sizes and should be accounted for during evaluation.
Another thing to consider is a relatively high 
OOV\footnote{We apply the concept of OOV not to words, but to lexical forms. Thus, if we have observed \textit{cut} only as a verb during training and it appears as a noun at test time, it is counted as unseen, even though it is the same \textit{cut} on the surface.} rate:
with 1/6 of test data unseen, overall performance strongly depends on success in handling OOV.
For this reason, whenever possible, in addition to overall performance we also report OOV results.
%
Lastly, as a rough morphological complexity estimate, we gather statistics on the average number of grammemes per lexical form.
The metric is used as a more convenient substitute to the degree of synthesis, defined as an average morpheme per word ratio
\cite{Greenberg1960}, which we have no reliable way of estimating, since the data is not annotated for morpheme boundaries\footnote{For instance, \textit{workers} consists of three morphemes (\textit{work-er-s}), but is annotated as bearing two grammemes and no morpheme boundaries, i.e. [\textit{worker}; n, pl].}.
For our data set the average grammeme-form ratio of 2.6 is about the same across all of the splits and is relatively high, attesting to the difficulty of the task.
For reference: the average grammeme-form ratio computed over five English treebanks is 1.9$\pm$0.05.


\subsection{\label{BSL} Parameter Settings and Baselines}

\begin{table}[!t]
\begin{center}
\small
\begin{tabular}{lrrr}
\hline

\bf Metric &  & \bf Splits &  \\
 & train & dev & test \\
\hline

sent-s, thou. & 6.1$\pm$9.2 & 0.8$\pm$1.2 & 0.8$\pm$1.2 \\
tokens, thou. & 108.1$\pm$165.8 & 13.5$\pm$20.6 & 13.5$\pm$20.5 \\
OOV rate & -- & 0.17$\pm$0.11 & 0.17$\pm$0.11 \\
gramm-form & 2.64$\pm$0.91 & 2.65$\pm$0.91 & 2.64$\pm$0.90 \\
\hline

\end{tabular}
\end{center}
\caption{\label{tab_data} Data set statistics: values correspond to means and standard deviations of respective metrics computed across all of the treebanks; here ``gramm-form" is an average number of grammemes per lexical form.}
\end{table}

LemMED is implemented in OpenNMT-py~\cite{opennmt}, v.0.5.0\footnote{\url{https://github.com/OpenNMT/OpenNMT-py/commit/b085d57}}.
Table~\ref{tab_params} lists the the key parameter-value pairs, which we use in all of the experiments, unless stated otherwise.
For optimization we use SGD with an initial learning rate of 1.0, which is halved every 10k training steps starting from the step \#25k.
OpenNMT-py v.0.5.0 does not implement a stopping criterion and training continues for a fixed number of steps, which we set to 50k.
Every 1000 steps the current version of the model is saved.
As a result we end up with 50 models, of which the best one is selected based on the dev set performance.
This model is then used for evaluation.

We compare LemMED to three baselines:
(i) SIGMORPHON~\cite{neubsl}, Shared task baseline, a pipeline of an LSTM tagger and seq2seq lemmatizer;
(ii) UDPipe~\cite{STRAKA16.873}, a statistical disambiguation model;
(iii) Lemming~\cite{Lemming}, a statistical joint learning model.
For SIGMORPHON we rely on results reported in Shared task review paper~\cite{sgm2019}.
For the other two baselines we use off-the-shelf implementations: UDPipe v.1.2.0 and Lemmingatize\footnote{\url{https://tinyurl.com/y2l7e6at}}, a python wrapper over the latest version of Lemming.
We use both systems with default settings, changing only those options that affect general task-related behaviour, e.g. instructing Lemming to use morpho-tags, which it ignores by default.
Both systems are trained for 50 iterations.
Similar to LemMED, UDPipe selects the best version of the model based on the dev set performance.
To our knowledge, Lemming does not employ any model selection strategy and uses the parameters updated and learnt during the final iteration of training.

\begin{table}[!t]
\begin{center}
\small
\begin{tabular}{lrr}
\hline

\bf Parameter &  \bf Value &  \bf Comment \\
\hline

encoder type & brnn & biLSTM \\
decoder type & rnn  & LSTM \\
\# layers & 2 & both enc./dec. \\
\# hidden units & 500 & both enc./dec. \\
dropout probability & 0.3 & both enc./dec. \\
attention type & global & \citet{global_att} \\
embeddings size & 700 & both source/target \\
beam size, at inference & 5 & default value \\

\hline
\end{tabular}
\end{center}
\caption{\label{tab_params} LemMED: values of the key parameters.}
\end{table}

Lastly, we would like to comment on the issues with the data format.
As already mentioned (cf. subsection~\ref{DATA}), the data was obtained by converting UD treebanks to the UniMorph scheme.
As a result, POS- and morpho-tags that were stored in separate CoNLL-U fields before conversion, end up being mixed in one field following no particular order.
This may potentially cause trouble for LemMED, therefore, for our model we preprocess the data by ordering grammemes alphabetically.
The baselines are also affected by formatting.
While UDPipe requires POS-tags and grammemes to be clearly separated, Lemming benefits from such an arrangement, albeit does not obligate it.
Nevertheless, for both baselines we convert the data back to the original format, using the UM-UD compatibility\footnote{\url{https://tinyurl.com/y4qft2cy}} mapping in reverse.





\subsection{\label{EVAL} Evaluation Metrics}

Arguably, the metric that matters the most for the task at hand is the analysis accuracy, i.e. the percentage of tokens for which both lemma and tag were correctly predicted.
Nevertheless, this metric was not considered in the shared task and the script provided by the organizers evaluates lemmatization and tagging performance separately, though on two levels.
On a coarse level simple accuracy is computed as percent correct.
On a finer level lemmatization is assessed in terms of average lemmatization distance, i.e. Levenshtein distance between predicted and golden lemmata.
For tagging a fine-grained metric is an average F1 score between predicted and golden tags\footnote{For example, if the predicted tag is $p=\{n\}$ and the golden tag is $g = \{n, pl\}$, then precision is $P=|p \cap g|/|p|=1$, recall is $R=|p \cap g|/|g|=0.5$, and F1 score is $F1 = 2PR / (P + R) = 0.67$.}.
We modify the script to include the calculation of analysis accuracy and consider five evaluation metrics altogether:
(i) lemmatization accuracy;
(ii)  average lemmatization distance;
(iii)  tagging accuracy;
(iv)  average tagging F1 score;
(v)  analysis accuracy.


\section{\label{RESDISC} Results and Discussion}

\begin{table}[!t]
\begin{center}
\small
\begin{tabular}{lrr}
\hline

\bf Model &  \bf Hardware &  \bf Quantity \\
\hline

LemMED & GPU, GeForce GTX, 1080 Ti & 1 \\
UDPipe & CPU, AMD Operton 6348 & 1 \\
Lemming & CPU, AMD Operton 6348 & 48 \\

\hline
\end{tabular}
\end{center}
\caption{\label{tab_hardware} Hardware used to train the models.}
\end{table}

In this section we discuss the results of three experiments designed to analyze the role of context in sequence-to-sequence morphological analysis.
First, in subsection~\ref{FULLSEQ} we try to answer whether it is better to analyze a sentence as one long sequence or break it down into smaller sub-sequences, and if the latter is preferable, what optimal length such sub-sequences should be.
Then, in subsection~\ref{TRN_COMPLEX} we investigate if size or linguistic nature of the data affect the choice of optimal context size.
Having decided on the size of context, in subsection~\ref{TGTSEQ} we investigate what type of context should be modeled within a target sequence, e.g. lemmata, morpho-tags, or a combination of both.
Finally, in subsection~\ref{FINAL_RES} we evaluate the best version of LemMED on the test set.

\begin{table*}[!t]
\setlength\fboxsep{0pt}
\begin{center}
\small
\begin{tabular}{l|rrrrr|r}
\hline
\bf Method & \bf LACC & \bf LDIS & \bf TACC & \bf T-F1 & \bf ACC & \bf TIME \\

\hline
\textbf{Full sequence} & & & & & & \\

LemMED & 
93.1 (80.9) & 
0.25 (0.62) & 
89.3 (\colorbox[gray]{0.79}{72.4}) & 
93.0 (84.6) & 87.9 (\colorbox[gray]{0.79}{66.5}) & 
1191.8$\pm$733.6 \\

SIGMORPHON & 
\colorbox[gray]{0.79}{96.7} (\colorbox[gray]{0.79}{83.7}) & 
\colorbox[gray]{0.79}{0.08} (\colorbox[gray]{0.79}{0.38}) & 
74.3 (65.8) & 
89.6 (82.4) & 
73.0 (58.9) & 
- \\

UDPipe & 
93.1 (70.4) & 
0.14 (0.65) & 
90.6 (63.9) & 
94.6 (80.5) & 
87.4 (51.8) & 
\colorbox[gray]{0.79}{37.1$\pm$18.4} \\


Lemming & 
94.9 (79.9) & 
0.10 (0.43) & 
\colorbox[gray]{0.79}{91.5} (70.8) & 
\colorbox[gray]{0.79}{95.3} (\colorbox[gray]{0.79}{85.5}) & 
\colorbox[gray]{0.79}{88.9} (61.9) & 
42.1$\pm$52.2 \\

\hline
\textbf{Context window} & & & & & \\

LemMED-CW0 & 94.8 (84.0) & 0.11 (0.35) & 79.0 (62.6) & 87.1 (79.5) & 77.3 (58.3) & 54.6$\pm$13.8 \\
LemMED-CW1 & \textbf{97.3} (\textbf{85.5}) & \textbf{0.06} (\textbf{0.31}) & 92.1 (72.9) & 95.3 (85.8) & 91.2 (67.8) & 122.9$\pm$27.4 \\
LemMED-CW2 & 97.1 (85.2) & \textbf{0.06} (0.32) & \textbf{92.5} (\textbf{74.6}) & \textbf{95.5} (\textbf{86.4}) & \textbf{91.5} (\textbf{68.9}) & 231.4$\pm$66.2 \\
LemMED-CW3 & 96.7 (84.8) & 0.08 (0.35) & 92.0 (74.1) & 95.2 (86.2) & 91.0 (68.6) & 281.1$\pm$101.6 \\

UDPipe-CW0 & 89.0 (71.3) & 0.26 (0.63) & 70.3 (56.5) & 81.5 (75.5) & 67.5 (47.0) & \textbf{4.6$\pm$2.2} \\
UDPipe-CW1 & 93.2 (70.6) & 0.13 (0.64) & 90.2 (62.4) & 94.5 (79.8) & 86.9 (50.5) & 52.9$\pm$27.0 \\
UDPipe-CW2 & 93.2 (70.6) & 0.13 (0.64) & 90.5 (63.2) & 94.6 (80.2) & 87.3 (51.2) & 102.6$\pm$54.4 \\
UDPipe-CW3 & 93.2 (70.5) & 0.13 (0.65) & 90.6 (63.3) & 94.6 (80.1) & 87.4 (51.1) & 143.1$\pm$74.0 \\


Lemming-CW0 & 89.1 (77.2) & 0.21 (0.49) & 68.9 (58.4) & 80.2 (77.6) & 66.0 (51.0) & 23.5$\pm$42.4 \\
Lemming-CW1 & 94.8 (79.5) & 0.10 (0.44) & 91.0 (67.2) & 95.2 (83.8) & 88.4 (58.8) & 54.1$\pm$55.1 \\
Lemming-CW2 & 94.8 (79.1) & 0.10 (0.45) & 91.1 (66.8) & 95.1 (83.6) & 88.5 (58.4) & 97.9$\pm$77.9 \\
Lemming-CW3 & 94.8 (79.0) & 0.10 (0.45) & 91.0 (66.3) & 95.1 (83.1) & 88.4 (58.0) & 109.5$\pm$98.6 \\

\hline
\end{tabular}
\end{center}
\caption{\label{RES-10} Performance on a sample of 10 treebanks in the full sequence and the context window modes. Here:
CW\textit{N} - context window of size \textit{N};
LACC - lemmatization accuracy;
LDIS - average lemmatization distance;
TACC - tagging accuracy;
T-F1 - average tagging F1 score;
ACC - analysis accuracy;
TIME - training time in minutes.
Results for unseen data are given in parentheses. Best results for the full sequence mode are highlighted; best results overall are shown in bold.}
\end{table*}

\subsection{\label{FULLSEQ} Full Sequence vs Context Window Mode}

In this experiment we compare two modes of operation of our model: full sequence (cf. Figure~\ref{EXFIG-LemMED-FS}) and context window (cf. Figure~\ref{EXFIG-LemMED-CW}).
Recall that using a model in either mode simply means training and evaluating the model on either complete sentences or overlapping snippets of a given fixed size.
This means that we can use the baselines, namely Lemming and UDPipe\footnote{For SIGMORPHON we report Shared task results, which are only available for the full sequence mode.}, in the context window mode as well.
We do not expect these models to be affected by context restriction, as both of them utilize local features, which is, in effect, a context window approach implemented internally.
For this experiment we chose 10 treebanks to produce a typologically-balanced sample of languages: three analytic (Afrikaans, Chinese, English); two nonconcatenative (Arabic, Hebrew); two fusional (Bulgarian, Russian); and three agglutinative (Hungarian, Finnish, Turkish).
In addition to performance, we also measure efficiency in terms of wall time spent on training.
Table~\ref{tab_hardware} lists the hardware that we use to train the models.

Table~\ref{RES-10} contains two sets of results pertaining to the models' performance and efficiency in the two modes.
As it can be seen, in the full sequence mode LemMED performs poorly achieving 93.1\%, 89.3\%, and 93.0\% in lemmatization, and tagging accuracy and F1 score respectively.
This corresponds to 3.6\%, 2.2\%, and 2.3\% net decrease compared to the best performing baselines and ranks our model only third on the respective metrics.
As for lemmatization distance, LemMED is ranked last, suggesting that, when it gets lemmata wrong, it is off by a greater number of edit operations, on average, than that of the baselines.

Despite poor performance on individual metrics, in terms of combined lemmatization and tagging (ACC), LemMED ranks second.
Another strength of LemMED appears to be in handling unseen data, performance on which is given in parentheses for each metric.
Here our model achieves the highest tagging and analysis accuracy of 72.4\% and 66.5\% with respective net improvement of 1.6\% and 4.6\% over the second best model.
On other metrics, with the exception of LDIS, LemMED is ranked second on unseen data.
Nevertheless, these promising aspects of performance are tainted by terrible inefficiency of the full sequence mode, where, on average, our model takes almost 20 hours to train on a single treebank.
As the time complexity of our model is polynomial with respect to the lengths of source and target sequences, the context window mode, where said sequences are much shorter, should speed things up.


For the context window mode we experiment with windows of size 0 (no context) to 3.
The first thing to notice is the fact that using any amount of context is superior to using no context at all.
This is observed across all models and metrics, except for LemMED, which does better in lemmatization accuracy and distance without context than with unlimited context (the full sequence mode).
This is where the advantages end, as on the remaining metrics, LemMED-CW0 loses 6-10\% to the full sequence version.
Thus, we do not consider the no context scenario in the rest of the experiments.
Also, as we expected, UDPipe and Lemming do not benefit from context restriction and perform on par or slightly worse than they do in the full sequence mode.
Moreover, these models are more efficient in the full sequence mode; hence we use them as such in all the experiments that follow.

\begin{figure*}[!t]
\includegraphics[width=16cm]{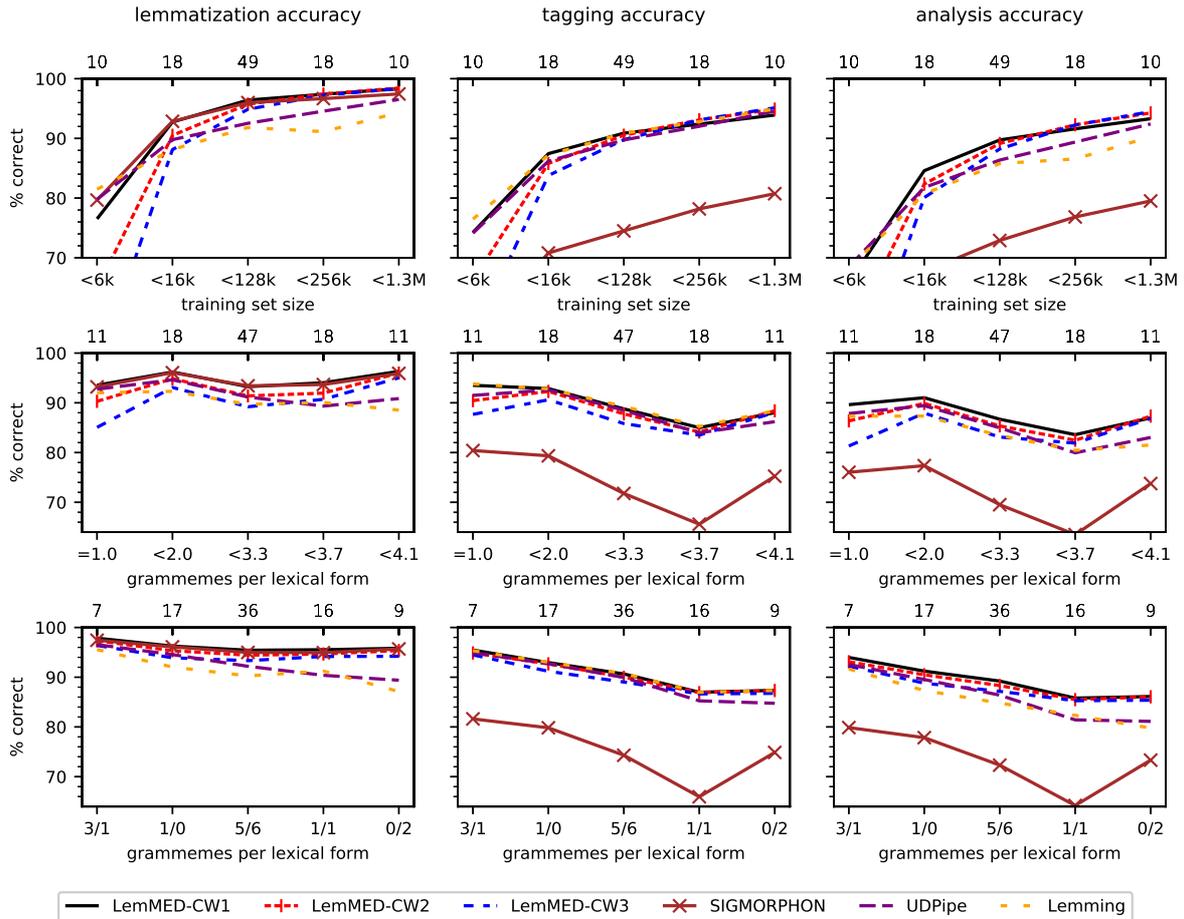}
\caption{\label{fig_tsize} The effect that size and morpho-complexity of the data have on performance.
The intervals are chosen in a way to achieve normal-like treebank per bin distributions, which are shown on the top horizontal axes.
The bottom row contains the complexity vs performance plots with 10 of the smallest and the largest treebanks removed.
Here the bottom horizontal axis shows the number of excluded treebanks in the smallest/largest format.}
\end{figure*}

In contrast to the baselines, our model shows clear improvement across all of the metrics when restricted to context windows of size 1 to 3.
The net increase in performance over the full sequence mode ranges from 2.2\% in tagging F-score (for LemMED-CW3) to 0.19 points (76\% relative) in lemmatization distance (for LemMED-CW1 and -CW2).
Moreover, training time reduces as context gets shorter and for LemMED-CW1 reaches a reasonable average of 2 hours per treebank.
Thus, limiting context to shorter sequences improves character-based sequence-to-sequence morphological analysis with respect to both performance and efficiency.
In fact, 
LemMED-CW2 achieves the best overall results on most of the metrics, improving 2.6\% in the analysis accuracy over the best performing baseline.
On the other hand LemMED-CW1 is only slightly less accurate, while being twice as fast to train.
Thus, before deciding on the optimal context size, we would like to compare all three models on a larger sample with more variation in size and linguistic properties of the data.


\subsection{\label{TRN_COMPLEX} Size and Morphological Complexity of the Data}

In order to investigate the influence that size and morpho-complexity of the data have on performance, we evaluate our models and the baselines on the dev sets of all the treebanks.
Then, we take the training sets and build distributions over two statistics (averaged per treebank): number of tokens and number of grammemes per lexical form.
The latter is used as a rough estimate of morphological complexity as explained in subsection~\ref{DATA}.
Lastly, we break each distribution into five bins and calculate average (per bin) performance for each model and baseline.

\begin{figure*}[!t]
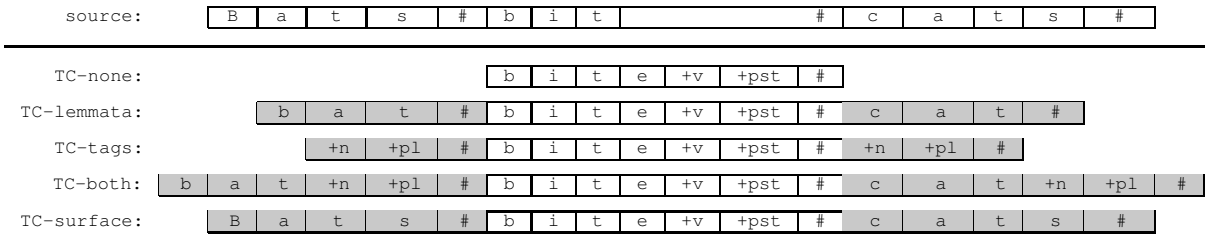

\center
\scriptsize
\begin{tabular}{r|c|c|c|c|c|c|c|c|c|c|c|c|c|c|c|c|c|c|c|l}

\cline{3-19}
\multicolumn{1}{r}{\texttt{source:}} & 
\multicolumn{1}{r|}{~} & 
\texttt{B} & \texttt{a} & \texttt{t} & 
\texttt{s} & \texttt{\#} & 

\texttt{b} & \texttt{i} & \texttt{t} & 
\multicolumn{4}{r|}{\texttt{\#}} & 

\texttt{c} & \texttt{a} & \texttt{t} & 
\texttt{s} & \texttt{\#} \\
\cline{3-19}
\multicolumn{20}{c} ~ \vspace{-1pt} \\
\hline
\multicolumn{20}{c} ~ \vspace{-1pt} \\

\cline{8-14}
\multicolumn{1}{r}{\texttt{TC-none:}} & 
\multicolumn{6}{r|}{~} &
\texttt{b} & \texttt{i} & \texttt{t} & \texttt{e} & 
\texttt{+v} & \texttt{+pst} & \texttt{\#} \\
\cline{8-14}
\multicolumn{20}{c} ~ \vspace{-2.3pt} \\

\cline{4-18}
\multicolumn{1}{r}{\texttt{TC-lemmata:}} & 
\multicolumn{2}{r|}{~} &
\cellcolor[gray]{0.79} \texttt{b} & 
\cellcolor[gray]{0.79} \texttt{a} & 
\cellcolor[gray]{0.79} \texttt{t} & 
\cellcolor[gray]{0.79} \texttt{\#} & 

\texttt{b} & \texttt{i} & \texttt{t} & \texttt{e} & 
\texttt{+v} & \texttt{+pst} & \texttt{\#} & 

\cellcolor[gray]{0.79} \texttt{c} & 
\cellcolor[gray]{0.79} \texttt{a} & 
\cellcolor[gray]{0.79} \texttt{t} & 
\cellcolor[gray]{0.79} \texttt{\#} \\
\cline{4-18}
\multicolumn{20}{c} ~ \vspace{-2.3pt} \\

\cline{5-17}
\multicolumn{1}{r}{\texttt{TC-tags:}} & 
\multicolumn{3}{r|}{~} &
\cellcolor[gray]{0.79} \texttt{+n} & 
\cellcolor[gray]{0.79} \texttt{+pl} & 
\cellcolor[gray]{0.79} \texttt{\#} & 

\texttt{b} & \texttt{i} & \texttt{t} & \texttt{e} & 
\texttt{+v} & \texttt{+pst} & \texttt{\#} & 

\cellcolor[gray]{0.79} \texttt{+n} & 
\cellcolor[gray]{0.79} \texttt{+pl} & 
\cellcolor[gray]{0.79} \texttt{\#} \\
\cline{5-17}
\multicolumn{20}{c} ~ \vspace{-2.3pt} \\

\cline{2-20}
\texttt{TC-both:} & 
\cellcolor[gray]{0.79} \texttt{b} & 
\cellcolor[gray]{0.79} \texttt{a} & 
\cellcolor[gray]{0.79} \texttt{t} & 
\cellcolor[gray]{0.79} \texttt{+n} & 
\cellcolor[gray]{0.79} \texttt{+pl} & 
\cellcolor[gray]{0.79} \texttt{\#} & 

\texttt{b} & \texttt{i} & \texttt{t} & \texttt{e} & 
\texttt{+v} & \texttt{+pst} & \texttt{\#} & 

\cellcolor[gray]{0.79} \texttt{c} & 
\cellcolor[gray]{0.79} \texttt{a} & 
\cellcolor[gray]{0.79} \texttt{t} & 
\cellcolor[gray]{0.79} \texttt{+n} & 
\cellcolor[gray]{0.79} \texttt{+pl} & 
\cellcolor[gray]{0.79} \texttt{\#} \\
\cline{2-20}
\multicolumn{20}{c} ~ \vspace{-2.3pt} \\

\cline{3-19}
\multicolumn{1}{r}{\texttt{TC-surface:}} & 
\multicolumn{1}{r|}{~} & 
\cellcolor[gray]{0.79} \texttt{B} & 
\cellcolor[gray]{0.79} \texttt{a} & 
\cellcolor[gray]{0.79} \texttt{t} & 
\cellcolor[gray]{0.79} \texttt{s} & 
\cellcolor[gray]{0.79} \texttt{\#} & 

\texttt{b} & \texttt{i} & \texttt{t} & \texttt{e} & 
\texttt{+v} & \texttt{+pst} & \texttt{\#} & 

\cellcolor[gray]{0.79} \texttt{c} & 
\cellcolor[gray]{0.79} \texttt{a} & 
\cellcolor[gray]{0.79} \texttt{t} & 
\cellcolor[gray]{0.79} \texttt{s} & 
\cellcolor[gray]{0.79} \texttt{\#} \\
\cline{3-19}

\end{tabular}
\caption{Possible target sequences for a given source sequence. Target-side context is highlighted if present.}
\label{EXFIG-TGTVAR}
\end{figure*}

Figure~\ref{fig_tsize} shows the resulting plots, where rows correspond to the data statistics and columns to the evaluation metrics.
Unsurprisingly, increase in training set size leads to increase in performance across almost\footnote{Except for lemmatization accuracy of Lemming, which drops on the interval of [128k, 256k) training tokens.} 
all models and metrics.
One would expect morpho-complexity to have the opposite effect, i.e. steady decline in performance with the growth of complexity.
However, according to Figure~\ref{fig_tsize} (middle row), performance of the models fluctuates on most of the metrics, typically rising on the first and the last intervals and falling in between.
Apparently, if ranked by the training set size, 3 of the bottom 10 and 2 of the top 10 treebanks fall into the first and the last bins respectively, with the bulk of the rest falling into the middle bin.
As we already know, all of the models perform at the opposite extremes on those two groups of treebanks.
If, for the purpose of our analysis, we consider those treebanks to be outliers and exclude them, we get the expected steady decline 
in tagging and analysis accuracy for all of the models, except SIGMORPHON (cf. Figure~\ref{fig_tsize}, bottom row).
As for lemmatization accuracy, with the outliers removed, the growth of morpho-complexity influences the statistical models to a greater degree and in a more expected fashion than it does the neural models.
This may suggest greater reliance of the statistical models on the tagging component.
LemMED, on the other hand, lacks a separate tagging component altogether, and, as we show in the next subsection, presence or absence of morpho-tags in the local context does not seem to influence lemmatization accuracy of our model.

When it comes to the baselines, none seem to cope with increasing morpho-complexity better than LemMED-CW1, performing worse than or on par with our model regardless of the evaluation metric.
What matters, though, is the data size, as all of the baselines beat our model on the smallest treebanks in every evaluation metric (ignoring SIGMORPHON's poor tagging and analysis performance).
Nevertheless, past the threshold of 6k training tokens our model delivers competitive performance in lemmatization and tagging accuracy, while clearly dominating the baselines in terms of analysis accuracy.

Lastly, we would like to comment on performance of LemMED relative to varying context size.
Intuitively, models with larger context windows capture sparser patterns and therefore require more training data to achieve best performance.
Indeed, LemMED-CW2 and -CW3 perform poorly on smaller treebanks, but quickly catch up with LemMED-CW1 as more data becomes available.
When there are 256k+ training tokens, the models with larger context windows even gain about 2\% in tagging and analysis accuracy over LemMED-CW1.
Also, single-word windows may not provide enough context to handle morphologically complex data.
This is evident from how fast the models that utilize more context close the performance gap with LemMED-CW1 (cf. Figure~\ref{fig_tsize}, middle row).
Even when the outliers are removed, performance of those models drops slower than that of LemMED-CW1, suggesting better resilience to increase in morpho-complexity.
Thus, in a data rich, morphologically complex scenario, where efficiency is not a concern, using context windows of two or three words may result in 1-2\% performance gain over LemMED-CW1.
However, in a scenario similar to ours, with many treebanks of varying size and morpho-complexity, using a single word of surrounding context is advisable for efficiency alone.
At any rate, for the data set at hand, LemMED-CW1 is clearly the strongest of our models; therefore in the subsequent experiments we use LemMED only in the single-word context window mode.

\subsection{\label{TGTSEQ} Target-side Context Variations}


So far, while training LemMED, we have been modeling target-side context by including both a lemma and a tag into each target side analysis.
However, as was noted in Section~\ref{MODEL}, unless voting on majority prediction is implemented, target side context is ignored at test time; and therefore can be arbitrary modified.
Figure~\ref{EXFIG-TGTVAR} depicts some of the possible modifications.
An obvious one is to do away with target-side context completely (TC-none), so that the model learns to predict only what matters and nothing else.
On the other hand, target-side context may not be completely redundant, as it may provide contextual clues during training.
Moreover, it might be the case that using only lemmata (TC-lemmata) or tags (TC-tags) tilts performance towards lemmatization or tagging respectively, while TC-both (i.e. current usage) maximizes analysis accuracy.
Lastly, for something that seems truly or, at least, more redundant than lexical context a blatant duplication of surface context can be used.
This variation (TC-surface) results in identical surface forms appearing in both source- and target-side context.


To check our intuitions, we need to train LemMED-CW1 using all of the target-side context types depicted on Figure~\ref{EXFIG-TGTVAR}, except ``target-both", for which the results are available from the previous experiments.
Doing so on the entire data set would mean training 420 models (four per treebank), which is extremely time consuming and resource demanding.
Instead, we train and test the models on the 10 treebanks sample that we used in subsection~\ref{FULLSEQ}.
The results are given in Table~\ref{RES-TGT}.
As can be seen, using any type of target-side context is better than using none at all.
Indeed, TC-none is beaten by every other variation, including TC-surface, in almost every metric.
Also, using TC-lemmata and TC-tags does tilt performance towards lemmatization and tagging respectively, yet not enough to outperform TC-both on those metrics (though, there are gains in OOV accuracy in both cases).
Overall, modeling complete target-side context, i.e. using both lemmata and tags, seems to be most beneficial, as TC-both beats or matches every other variation on every metric.

\begin{table}[!t]
\begin{center}
\small
\begin{tabular}{lrrr}
\hline
\textbf{TC type} & \textbf{LACC} & \textbf{TACC} & \textbf{ACC} \\
\hline


None & 96.5 (84.8) & 91.1 (72.9) & 90.2 (67.6) \\

Lemmata & \textbf{97.3} (\textbf{85.8}) & 91.0 (72.6) & 90.2 (67.6) \\

Tags & 96.6 (85.0) & \textbf{92.1} (\textbf{73.2}) & 90.5 (67.0) \\

Both & \textbf{97.3} (85.5) & \textbf{92.1} (72.9) & \textbf{91.2} (\textbf{67.8}) \\

Surface & 96.9 (85.1) & 91.3 (72.9) & 90.4 (67.5) \\


\hline
\end{tabular}
\end{center}
\caption{\label{RES-TGT} Performance of LemMED-CW1 with respect to the types of target-side context.
The results are averaged over the 10 treebank sample.}
\end{table}

\subsection{\label{FINAL_RES} Test Set Performance}

For the final test set evaluation we use LemMED-CW1 with complete target-side context modeling.
In addition, we implement voting on majority prediction and refer to this version of the model as LemMED-V.
Bearing in mind sensitivity of our model to data size, we also consider a separate scenario that excludes 10 smallest treebanks, i.e. the treebanks with 6k or less training tokens.
The results are shown in Table~\ref{RES-FIN}.

First, we notice that LemMED-V achieves 0.1-0.3\% improvement over the basic model in general performance on all of the metrics, but loses 0.2\% and 0.1\% in OOV performance in tagging and analysis accuracy respectively.
Second, regardless of the presence or absence of the smallest treebanks, both versions of LemMED improve 2.4-2.9\% in analysis accuracy over the next best model (UDPipe).
Third, on the complete data set our model outperform SIGMORPHON in lemmatization only if used in the voting mode.
However, when the smallest treebanks are removed LemMED-CW1 beats this baseline and, when it comes to OOV lemmatization accuracy, rather decisively.
Fourth, Lemming proofs to be a very strong tagging baseline, as only on the bigger treebanks can LemMED-V achieve a matching tagging accuracy, while still losing in tagging accuracy on OOV.
Overall, the conclusions we drew from a number of dev set evaluations are valid here as well.
While performing slightly better or slightly worse than the strongest baselines in lemmatization and tagging, LemMED is definitely better at predicting whole lexical forms, especially on unseen data.

\begin{table}[!t]
\begin{center}
\small
\setlength\fboxsep{0pt}

\begin{tabular}{lrrr}
\hline
\bf Method & \bf LACC & \bf TACC & \bf ACC \\

\hline
\multicolumn{4}{c}{\textbf{data, all}} \\
\hline

LemMED-V & \textbf{94.3} (\textbf{78.4}) & 89.0 (64.6) & \textbf{87.3} (57.3) \\

LemMED & 94.1 (78.3) & 88.8 (64.8) & 87.0 (\textbf{57.4}) \\
UDPipe & 91.5 (70.7) & 88.1 (61.0) & 84.6 (49.7) \\
Lemming & 90.3 (71.8) & \textbf{89.2} (\textbf{65.9}) & 83.6 (52.3) \\
SIGMORPH. & 94.2 (77.2) & 73.1 (62.0) & 70.7 (52.0) \\

\hline
\multicolumn{4}{c}{\textbf{data, 6k+ training tokens}} \\
\hline

LemMED-V & \colorbox[gray]{0.79}{96.0} (\colorbox[gray]{0.79}{82.4}) & \colorbox[gray]{0.79}{90.6} (67.7) & \colorbox[gray]{0.79}{89.2} (61.5) \\

LemMED & 95.9 (82.3) & 90.4 (67.7) & 89.0 (\colorbox[gray]{0.79}{61.6}) \\
UDPipe & 92.7 (71.6) & 89.6 (63.5) & 86.3 (51.9) \\
Lemming & 91.2 (72.8) & \colorbox[gray]{0.79}{90.6} (\colorbox[gray]{0.79}{68.1}) & 85.2 (54.3) \\
SIGMORPH. & 95.5 (80.5) & 74.9 (64.2) & 73.0 (55.6) \\

\hline
\end{tabular}
\end{center}
\caption{\label{RES-FIN} Final results on the test set.
LemMED-V denotes the version with voting on majority prediction.
Best results overall are shown in bold.
Best results on the treebanks with 6k+ training tokens are highlighted.}
\end{table}

\section{\label{ST19} Shared Task Performance}

Before concluding the paper we would like to report on our model's Shared task performance, where we have submitted the version of LemMED-CW1 used in subsection~\ref{FINAL_RES} (without voting on majority prediction).
Table~\ref{RES-ST19} compares the results of the top 10 systems ranked by the average value calculated over the four metrics used in the shared task.
To save space, system names are replaced by letters in alphabetic order.
A version of the table with the results of all of the submitted systems and their full names is available in the overview paper by~\citet{sgm2019}.

\begin{table}[!t]
\begin{center}
\small
\begin{tabular}{rlrrrr|r}
\hline
\textbf{\#} & \textbf{Sys.} & \textbf{LACC} & \textbf{LDIS} & \textbf{TACC} & \textbf{T-F1} & \textbf{AVG} \\
\hline

1 & A* & 95.78 & 0.10 & 93.19 & 95.92 & 93.76 \\
2 & B* & 95.00 & 0.11 & 93.23 & 96.02 & 93.37 \\
3 & C* & 93.91 & 0.14 & 90.53 & 94.54 & 91.30 \\
4 & D*+ & 93.94 & 0.15 & 90.61 & 94.57 & 90.98 \\
5 & \textbf{LM}\footnotemark & 94.20 & 0.13 & 88.93 & 92.89 & 90.64 \\
6 & E* & 94.07 & 0.13 & 88.09 & 91.84 & 90.33 \\
7 & F & 94.46 & 0.11 & 86.67 & 90.54 & 90.22 \\
8 & G & 93.06 & 0.15 & 88.80 & 93.22 & 89.94 \\
9 & H+ & 93.05 & 0.16 & 89.00 & 93.35 & 89.81 \\
10 & I & 93.43 & 0.17 & 87.42 & 92.51 & 89.14 \\

\hline
- & \textbf{SM} & 94.17 & 0.13 & 73.16 & 87.92 & 85.60 \\

\hline
\end{tabular}
\end{center}
\caption{\label{RES-ST19} Shared task performance.
The systems are ranked by descending average over the four metrics.
When computing the average LDIS is scaled as $100(1-LDIS)$.
Here:
LM denotes LemMED-CW1;
SM denotes SIGMORPHON;
``*" means that the system uses external data;
``+" means that the system was submitted past the deadline.}
\end{table}

Overall, LemMED is ranked fifth in terms of average performance, lemmatization, and tagging accuracy.
It is tied for fourth and fifth in lemmatization distance and is ranked seventh in tagging F1 score.
However, if we ignored the systems that use external resources, our model would rank second in the first three metrics, third in TF-1, and first in average performance.
Thus, for a straightforward model that it is, LemMED achieves a competitive performance among systems that do not utilize external resources.
Unfortunately, the organizers do not report analysis accuracy and results on unseen data, which would be interesting to compare, given that those are the two strongest aspects of LemMED's performance.

\section{\label{CON} Conclusion}

\footnotetext{The shared task results for our system differ from those reported in Table~\ref{RES-FIN}, because two treebanks were left out in the latter to allow for comparison with Lemming (cf. subsection~\ref{DATA}).}

We presented LemMED, a character-level encode-decoder with attention applied to contextual morphological analysis.
We showed how it extends similar attention-based models developed for lemmatization (Lematus) and non-contextual inflection (MED).
Upon investigating the role of context, we have concluded that using small context windows to model short sequences is preferable to modeling whole sentences.
It was found that a choice of specific window size may depend on characteristics of the data, such as size and morpho-complexity.
In our case using a single-word context window yielded best average performance, while maintaining an acceptable efficiency of two hours of training per treebank.
In addition to context size, we experimented with various types of target-side context and concluded that it should be modeled in full, i.e. including both lemmata and tags.
We evaluated LemMED on a set of more than a hundred treebanks, comparing it to both popular baselines available off-the-shelf and state of the art systems submitted to the SIGMORPHON-2019 shared task on contextual analysis.
In both cases our model delivered a competitive performance and when compared to the baselines showed clear dominance in terms of combined lemmatization and tagging accuracy and performance on unseen data.

\bibliography{ref}
\bibliographystyle{acl_natbib}

\end{document}